\documentclass[10pt,twocolumn,letterpaper]{article}

\usepackage[pagenumbers]{cvpr}      
\usepackage{tcolorbox}
\definecolor{dt}{gray}{0.7}
\usepackage{multicol}
\usepackage{multirow}
\usepackage{makecell}
\usepackage{color}
\usepackage{xcolor}
\usepackage{colortbl}
\usepackage[inkscapelatex=false]{svg}
\DeclareSymbolFont{extraup}{U}{zavm}{m}{n}
\DeclareMathSymbol{\varheart}{\mathalpha}{extraup}{86}
\DeclareMathSymbol{\vardiamond}{\mathalpha}{extraup}{87}
\definecolor{cvprblue}{rgb}{0.21,0.49,0.74}
\usepackage[pagebackref,breaklinks,colorlinks,citecolor=cvprblue]{hyperref}

\def\paperID{*****} 
\def\confName{CVPR}
\def\confYear{2024}

\title{Lyrics: Boosting Fine-grained Language-Vision Alignment via \\ Semantic-aware Visual Objects}
\author{
    \textbf{Junyu Lu}$^{\varheart}$\footnotemark[1] \qquad
    \textbf{Dixiang Zhang}$^{\varheart\vardiamond}$\footnotemark[1] \qquad
    \textbf{Songxin Zhang}$^{\clubsuit}$\footnotemark[1] \qquad
    \textbf{Zejian Xie}$^{\clubsuit}$ \qquad
    \textbf{Zhuoyang Song}$^{\clubsuit}$ \qquad
    \\
    \textbf{Cong Lin}$^{\clubsuit}$ \qquad
    \textbf{Jiaxing Zhang}$^{\varheart}$\footnotemark[2] \qquad
    \textbf{Bingyi Jing}$^{\clubsuit}$\footnotemark[2] \qquad
    \textbf{Pingjian Zhang}$^{\vardiamond}$\footnotemark[2] \qquad
    \\
    $^{\varheart}$International Digital Economy Academy \quad
    $^{\vardiamond}$South China University of Technology \quad \\
    $^{\clubsuit}$Southern University of Science and Technology \quad \\
    {\tt\small lujunyu@idea.edu.cn, zhangdixiang@mail.scut.edu.cn, zhangsx@mail.sustech.edu.cn} \\
    {\tt\small zhangpingjian@scut.edu.cn, jingby@sustech.edu.cn}
}

\begin{document}
\maketitle

{
  \renewcommand{\thefootnote}%
    {\fnsymbol{footnote}}
  \footnotetext[1]{Equal Contribution.}
  \footnotetext[2]{Corresponding Author.}
}

\begin{abstract}
Large Vision Language Models (LVLMs) have demonstrated impressive zero-shot capabilities in various vision-language dialogue scenarios. However, the absence of fine-grained visual object detection hinders the model from understanding the details of images, leading to irreparable visual hallucinations and factual errors. In this paper, we propose Lyrics, a novel multi-modal pre-training and instruction fine-tuning paradigm that bootstraps vision-language alignment from fine-grained cross-modal collaboration. Building on the foundation of BLIP-2, Lyrics infuses local visual features extracted from a visual refiner that includes image tagging, object detection and semantic segmentation modules into the Querying Transformer, while on the text side, the language inputs equip the boundary boxes and tags derived from the visual refiner. We further introduce a two-stage training scheme, in which the pre-training stage bridges the modality gap through explicit and comprehensive vision-language alignment targets. During the instruction fine-tuning stage, we introduce semantic-aware visual feature extraction, a crucial method that enables the model to extract informative features from concrete visual objects. Our approach achieves robust performance on 13 datasets across various vision-language tasks, and demonstrates promising multi-modal understanding, perception and conversation capabilities in 11 scenario-based benchmark toolkits.

\end{abstract}    
\section{Introduction}
\label{sec:intro}

Large language models (LLMs) have attracted widespread attention in the artificial intelligence community due to their powerful language generation and comprehension capabilities~\cite{brown2020gpt3,touvron2023llama,vicuna2023}. These models can perform a variety of intricate linguistic tasks by further learning user intentions in elaborate instruction tuning datasets~\cite{wei2021flan}. To explore the potential of LLMs beyond language, recent studies develop the large-scale vision-language models (LVLMs) to perceive and understand visual signals while inheriting advanced logical reasoning and knowledge generalizing capabilities from LLMs~\cite{alayrac2022flamingo,liu2023llava,li2023blip2,dai2024instructblip,chen2023shikra}. With unified format vision-language instructions and proper visual perceiver, prominent LVLMs demonstrate impressive performance in detailed image description, referential dialogues and complex multi-modal reasoning under real-world scenario.

However, widely-used LVLMs habitually adopt Vision Transformer (ViT)~\cite{dosovitskiy2020vit} from pre-trained CLIP~\cite{radford2021clip} as the image encoder, whose visual feature generalization capabilities are manifested in executing pre-defined label classification and brief image-text matching. Therefore, learning to effectively detect fine-grained visual objects within images (e.g. color, count, detailed description) and capturing visual morphology (e.g. action recognition, localization) present considerable challenges due to the lack of precise local visual features. Therefore, in situations where the image encoder fails to provide sufficient visual signals to meet the requirements of the specific objectives mentioned in the instructions, the LVLMs tends to produce incorrect responses that deviate from the details of the image. As analysed in Section~\ref{sec:case}., in dialogues involving visual objects, such as ``How many people in the image?'' and ``What sport are the people playing?'', existing LVLMs are limited by visual signals and prone to generating visual hallucinations.

\begin{figure*}[t]
    \centering
    \includegraphics[width=1\linewidth, trim=0 10 0 0]{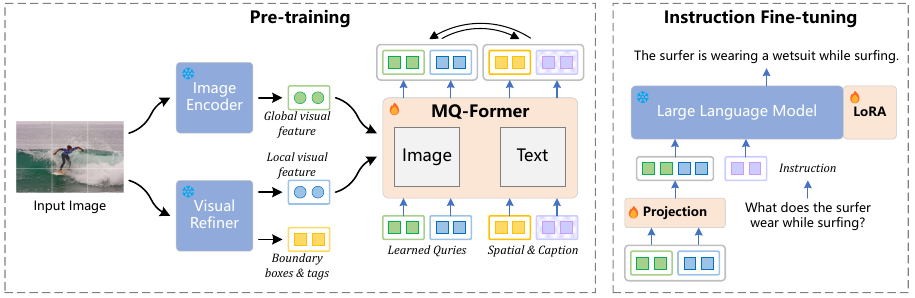}
    \vspace{-0.05in}
    \caption{The two-stage training framework of Lyrics, with the MQ-Former to bridge the modality gap between the image encoder and the visual refiner to the LLM. The first stage bootstraps vision-language representation alignment via multi-task pre-training. The second stage bootstraps instructed vision-language generative learning via semantic-aware visual objects.}
    \label{fig:two_stage_training}
    \vspace{-0.05in}
\end{figure*}

To prevent the deficiency of visual signals from hindering the expression of LLM, we propose Lyrics, a fine-grained vision-language pre-training and instruction fine-tuning framework that enables the model to handle semantic-aware visual objects as Figure~\ref{fig:two_stage_training}. Lyrics is initialized from a pre-trained BLIP-2~\cite{li2023blip2} model, which introduces Querying Transformer to align vision-language representations and bridge a LLM and an image encoder. For better visual perception, we construct a visual refiner that consist of an image tagging module~\cite{zhang2023ram}, an object detection module~\cite{zhang2022dino} and a semantic segmentation module~\cite{kirillov2023sam}. Specifically, the image tagging module can recognize any common categories. The object detection module and semantic segmentation module can further extract local visual features related to locating visual objects and generating semantic masks, which can be used to convert abstract visual signals into concrete spatial representation. We further introduce a Multi-scale Querying Transformer (MQ-Former), which takes local visual features and concrete spatial representation provided by the visual refiner to bootstrap vision-language alignment. In the pre-training stage for vision-language representation alignment, we employ a pair of learnable query vectors to compress both the local visual features from the visual refiner and the global visual features from the image encoder. We utilize the boundary boxes and tags of visual objects decoded from the visual refiner, together with the image caption and learnable queries, to perform various semantic alignment tasks. In the instruction fine-tuning stage, we connect the learned queries output from MQ-Former to the LLM for instruction-response generative learning, and train low-rank adaptation (LoRA)~\cite{hu2021lora} on the LLM. Our main contributions are summarized as:
\begin{itemize}
    \item We develop a novel vision-language alignment paradigm to explore the fine-grained relation between multi-scale visual and textual signals, which employs the local visual features and spatial representation extracted from the visual refiner for representation learning.
    \item We propose Lyrics, a generalist LVLM that understand and perceive semantic-aware visual objects via a two-stage training framework, for achieving precise visual knowledge understanding and reasoning capabilities.
    \item we conduct extensive experiments on diverse vision-language tasks, including image captioning, visual question answering (VQA) and referring expression comprehension (REC). The results demonstrate that Lyrics can achieve state-of-the-art or comparable performance on several benchmarks compared to previous LVLMs.
\end{itemize}

\section{Related Work}
\label{sec:related work}

\subsection{Advanced Large Language Models}

Early language models such as GPT-2~\cite{radford2019gpt2} and BERT~\cite{devlin2019bert} are foundation models trained on large-scale web-crawled datasets, symbolizing milestones in the NLP field for text understanding. Following the success of structures and training strategies, numerous LLMs showcase significant zero-shot text understanding and generation capabilities with the scaling up of training data and model size, such as GPT-3~\cite{brown2020gpt3}, PaLM~\cite{chowdhery2022palm} and BLOOM~\cite{scao2022bloom}. Consequently, the recent representative work, LLaMA~\cite{touvron2023llama}, focuses on refining LLMs to engage in human instruction and feedback. LLaMA is fine-tuned on high-quality instruction datasets, demonstrating powerful instruction-following and human interaction capabilities, which facilitates the continued training of various impressive works, such as Alpaca~\cite{alpaca}, Vicuna~\cite{vicuna2023} and MPT~\cite{MosaicML2023mpt}.

\subsection{Large Vision-Language Models}

With remarkable generalization and robustness of LLMs, common LVLMs use a vision-language cross-modal adapter to align the visual features from the visual encoder with the LLMs, thereby stimulating the ability of LLMs to perceive and understand visual signals. Flamingo~\cite{alayrac2022flamingo} freezes the pre-trained visual encoder and LLMs and integrates multi-modal representations through perceiver and gated cross-attention, demonstrating impressive few-shot capabilities. Meanwhile, BLIP-2~\cite{li2023blip2} trains a Q-Former to compress visual features as input to the frozen LLMs. On this basis, InstructBLIP~\cite{dai2024instructblip} proposes instruction-aware visual feature extraction that enables flexible and informative feature extraction according to the given instructions. Early work such as LLaVA~\cite{liu2023llava} and Mini-GPT4~\cite{zhu2023minigpt4} attempt to simply feed visual features into LLMs using only a learnable linear layer, which introduce visual instruction tuning to enhance instruction following capabilities in LVLMs. Furthermore, concurrent works such as Vision-LLM~\cite{wang2023visionllm}, Kosmos-2~\cite{peng2023kosmos2}, Shikra~\cite{chen2023shikra} and Qwen-VL~\cite{bai2023qwen} also demonstrate that the open training on visual encoders and LLMs can promote the LVLMs to understand located objects within the images and generate text formats of bounding boxes to perform visual grounding.

\section{Method}
\label{sec:method}

We propose Lyrics, a novel two-stage training scheme that bootstraps fine-grained vision-language alignment via semantic-aware visual objects: (1) The pre-training stage aligns multi-scale visual and textual features within MQ-Former. (2) The instruction fine-tuning stage connects the MQ-Former to the LLMs to perform semantic-aware vision-to-language generative learning. This section begins with an introduction to the model architecture of MQ-Former with visual refiner, followed by the delineation of fine-grained two-stage training scheme.

\begin{figure*}[t]
    \centering
    \includegraphics[width=1\linewidth, trim=0 10 0 0]{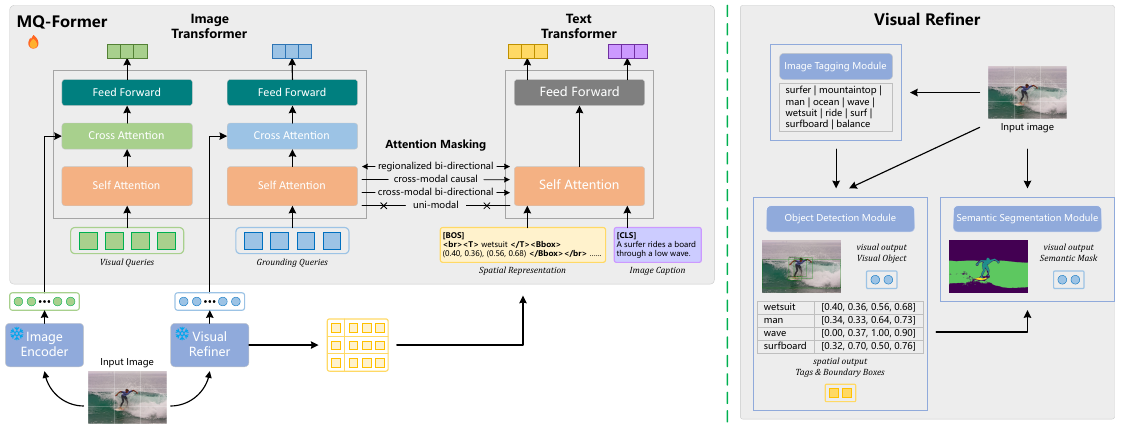}
    \vspace{-0.05in}
    \caption{\textit{(Left)} Model architecture of Multi-scale Querying Transformer (MQ-Former), The frozen global and local visual features are inserted into every image transformer block to interact with learnable quries. \textit{(Right)} The pipeline of visual refiner that consists of a image tagging module, an object detection module and a semantic segmentation module.}
    \label{fig:MQ-Former}
    \vspace{-0.1in}
\end{figure*}

\begin{figure}[t]
    \centering
    \includegraphics[width=1\linewidth, trim=0 10 0 0]{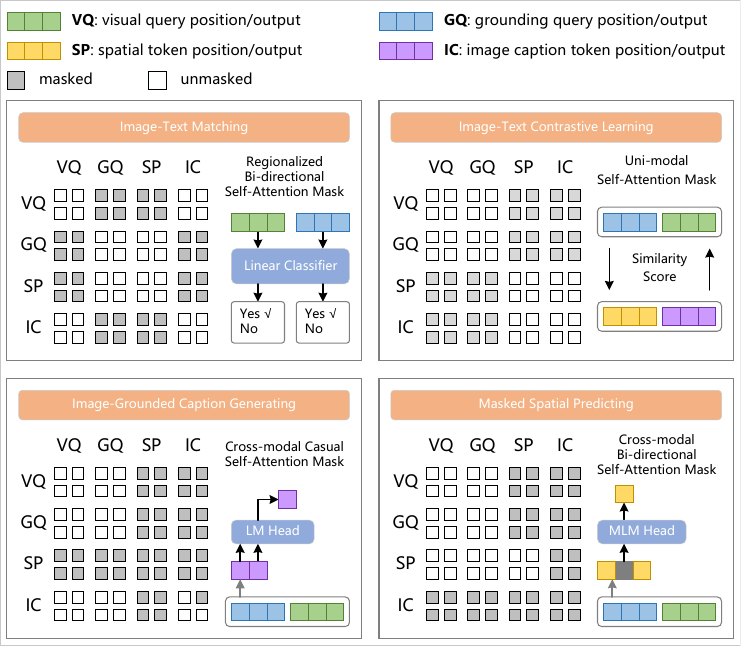}
    \caption{The learning objectives in vision-language representation alignment. We jointly optimize four objectives which enforce the queries (a set of learnable embeddings) to extract visual representation relevant to the text information. The self-attention masking strategy for each objective is used to control query-text interaction.}
    \label{fig:attention_mask}
    \vspace{-0.2in}
\end{figure}

\subsection{Model Architecture}

To excavate visual objects within images and establish correlation with spatial representation, we introduce a visual refiner composed of an image tagging module, an object detection module and a semantic segmentation module, as illustrated in Figure~\ref{fig:MQ-Former} \textbf{(Right)}. Concretely, for an given image, we first employ the Recognize Anything Model (RAM)~\cite{zhang2023ram}, a strong foundation model for zero-shot image tagging that incorporates semantic information into label queries, to generate any common categories relevant to the semantic object with the image. We denote the tag set with $N_t$ detected tags as $Tag = \left\{t_i\right\}_1^{N_t}$ and concatenate each tag into a sentence $t_1, t_2, \dots, t_{N_t} $ using comma. Then, we transmit the image and sentence to Grounding-DINO~\cite{zhang2022dino}, a open-set Transformer-based object detection model that performs vision-language modality fusion at multiple phases. For each tag, we obtain all boundary boxes beyond the filtering threshold from Grounding-DINO, and define the matched tags and boundary boxes as the spatial information $\{t_i: [x_i^1, y_i^1, x_i^2, y_i^2]\}$ for the $i$-th visual object. Additional, we feed the image and its spatial coordinates into the Segment Anything Model (SAM)~\cite{kirillov2023sam}, a lightweight image segmentation framework that can generate local visual features related to the semantic mask of visual objects. Formally, with the image tagging module and object detection module, we obtain all boundary boxes and tags and formulate them into the textualized format of $Spatial\ Rep =\langle\mathrm{br}\rangle\langle\mathrm{T}\rangle \ t_1\ \langle/ \mathrm{T}\rangle\langle\mathrm{Bbox}\rangle\left(x_i^1, y_i^1\right),\left(x_i^2,\ y_i^2\right)\langle/ \mathrm{Box}\rangle\langle/ \mathrm{br}\rangle$ as the spatial representation of semantic-aware visual objects. Furthermore, We concatenate the local visual features extracted from the object detection module and semantic segmentation module as the visual output of the visual refiner, and synchronously use a Vision Transformer (ViT)~\cite{dosovitskiy2020vit} as an image encoder to extract global visual features.

To bridge the modality gap between image encoder and visual refiner to LLMs, we propose MQ-Former as a trainable module to perform vision-language alignment. As shown in Figure~\ref{fig:MQ-Former} \textbf{(Left)}, MQ-Former consists of two transformer~\cite{vaswani2017transformer} submodules that share the same self-attention layer. (1) In the image transformer, we create a set of fixed-quantity visual quries and grounding quries, which interact with the image encoder and visual refiner respectively to output compressed visual features. It infuses the high-dimension visual features from the image encoder and visual refiner into the learnable quries through two independent cross-attention layers. The grounding quries and visual quries share a feed forward network equipped with a ReLU activation function for feature transformation. (2) The text transformer takes the concatenated spatial representation and image caption as input text, each prefixed with special tokens \texttt{[BOS]} and \texttt{[CLS]} at the outside. Additionally, we perform intra-modal and cross-modal interactions between queries and text representations in the self-attention layer and control information fusion through the attention mask. 

In our experiments, MQ-Former continues training on the BLIP-2~\cite{li2023blip2} first-stage pre-training breakpoints, and we employ Xavier initialization~\cite{glorot2010xiaver} to configure the extra cross-attention layer for grounding quries. we use 32 grounding quries and 32 visual quries, each with a dimension of 768, which is consistent with the hidden dimension of MQ-Former. In this way, the output query representation is much smaller than the size of frozen visual feature (e.g. $257\times1024$ for ViT-L/14 and $260\times900$ for Grounding-DINO-T).


\subsection{Bootstraping Vision-Language Representation Alignment via Multi-Task Pre-training}

During the pre-training stage, we connect MQ-Former to the frozen image encoder and visual refiner, controlling the mutual visibility of queries and text representations through self-attention mask matrix to perform various pre-training tasks. Refer to BLIP-2~\cite{li2023blip2}, we jointly optimize four objectives to pre-train MQ-Former as Figure~\ref{fig:attention_mask}, which influences the visual queries and grounding quries for extracting visual features that are more informative of the spatial and textual representations. Formally, to quantify the role of each query token, we separately pass the output embeddings of visual queries and grounding queries through a pooling layer, and concatenate the pooled outputs $H_v$ and $H_g$ as $H_I$. Similarly, We concatenate the output embeddings $H_{sp}$ and $H_{ic}$ of the \texttt{[BOS]} and \texttt{[CLS]} tokens to form $H_T$, which represent spatial information and image caption. Four losses are delineated as following.


\paragraph{\textbf{Image-Text Contrastive Learning}} (ITC) learns to align the fine-grained visual and text representations by encouraging positive image-text pairs to have similar representations in contrast to the negative pairs. we mutually mask the text and image transformers to avoid information leak, and calculate image-text similarity between the visual representation $H_I$ and the text representation $H_T$. We denote the softmax-normalized image-to-text and text-to-image similarity as $\boldsymbol{p}^{\mathrm{i} 2 \mathrm{t}}$ and $\boldsymbol{p}^{\mathrm{t} 2 \mathrm{i}}$, and the ground-truth
one-hot similarity as $\boldsymbol{y}^{\mathrm{i} 2 \mathrm{t}}$ and $\boldsymbol{y}^{\mathrm{t} 2 \mathrm{i}}$. The image-text contrastive loss is defined as the cross-entropy H between $\boldsymbol{p}$ and $\boldsymbol{y}$:
\begin{align}
\vspace{0.1in}
\mathcal{L}_{\mathrm{itc}} =&\frac{1}{2} \mathbb{E}_{(I, T)}\left[\mathrm{H}\left(\boldsymbol{y}^{\mathrm{i} 2 \mathrm{t}}(H_I), \boldsymbol{p}^{\mathrm{i} 2 \mathrm{t}}(H_I)\right) \right. \notag\\&+\left.\mathrm{H}\left(\boldsymbol{y}^{\mathrm{t} 2 \mathrm{i}}(H_T), \boldsymbol{p}^{\mathrm{t} 2 \mathrm{i}}(H_T)\right)\right]
\end{align}

\paragraph{\textbf{Image-Text Matching}} (ITM) is a binary classification task that transmits the pooled query embedding $H_v$ and $H_g$ into a classifier followed by softmax to predict a two-class probability $\boldsymbol{p}^{\mathrm{itm}}_v$ and $\boldsymbol{p}^{\mathrm{itm}}_g$. It aims to learn image-text representations to capture both coarse-grained and fine-grained semantic alignments between vision and language. We employ a regionalized bi-directional self-attention mask that permits mutual interaction between visual queries and image caption, as well as between grounding queries and spatial representation, while remaining tokens are prevented from attaching to each other. Let $\boldsymbol{y}^{\mathrm{itm}}_v$ and $\boldsymbol{y}^{\mathrm{itm}}_g$ denote 2-dimension one-hot vectors representing the ground-truth label. The ITM loss is:
\begin{align}
\mathcal{L}_{\mathrm{itm}}=&\frac{1}{2} \mathbb{E}_{(I, T)} \left[\mathrm{H}\left(\boldsymbol{y}^{\mathrm{itm}}_v, \boldsymbol{p}^{\mathrm{itm}}(H_v, H_{ic})\right) \right. \notag\\&\left. + \mathrm{H}\left(\boldsymbol{y}^{\mathrm{itm}}_g, \boldsymbol{p}^{\mathrm{itm}}(H_g, H_{sp})\right)\right]
\end{align}

\paragraph{\textbf{Image-Grounded Caption Generating}} (ICG) optimizes the MQ-Former to enable it to generate image caption solely based on visual features from visual quries and grounding quries. As the text tokens cannot directly interact with the image encoder and the visual refiner, ICG task enables the MQ-Former with multi-modal generalization capabilities to convert abstract visual feature into coherent image caption. Similar to UniLM~\cite{dong2019unilm}, we utilize a cross-modal causal self-attention mask to control query-caption interactions, with spatial information being masked accordingly. We also replace the \texttt{[CLS]} with \texttt{[DEC]} to signify language modeling task. Let $\boldsymbol{y}^{\mathrm{icg}}$ denote the masked image caption and $\boldsymbol{p}^{\mathrm{icg}}$ denote the predicted probability for a masked token $\hat{T}_{ic}$. ICG minimizes a cross-entropy loss:
\begin{equation}
\mathcal{L}_{\mathrm{icg}}=\mathbb{E}_{(I, \hat{T}_{ic})} \mathrm{H}\left(\boldsymbol{y}^{\mathrm{icg}}, \boldsymbol{p}^{\mathrm{icg}}(\left[H_v, H_g\right], \hat{T}_{ic})\right)
\end{equation}

\paragraph{\textbf{Masked Spatial Predicting}} (MSP) aims to learn semantic-aware visual objects through fine-grained multi-modal alignment. Referencing the whole word masking strategy, we adopt a 15\% probability to randomly replace all tokens with the spatial representation of a whole visual object with the special token \texttt{[MASK]}, requiring the model to restore the masked tags and boundary boxes via local visual features. We use a cross-modal bi-directional self-attention mask where intra-modal tokens are mutually visible and spatial representation can associate with queries, with the exception that image caption are entirely masked. We also replace the \texttt{[BOS]} with \texttt{[MLM]} to signify masked language modeling task. Let $\boldsymbol{y}^{\mathrm{msp}}$ denote the masked spatial representation and $\boldsymbol{p}^{\mathrm{msp}}$ denote the predicted probability for a masked token $\hat{T}_{sp}$. MSP minimizes a cross-entropy loss:
\begin{equation}
\mathcal{L}_{\mathrm{msp}}=\mathbb{E}_{(I, \hat{T}_{sp})} \mathrm{H}\left(\boldsymbol{y}^{\mathrm{msp}}, \boldsymbol{p}^{\mathrm{msp}}(\left[H_v, H_g\right], \hat{T}_{sp})\right)
\end{equation}

The full training objective of MQ-Former can be formulated as:
\begin{equation}
\mathcal{L}=\mathcal{L}_{\mathrm{itc}}+\mathcal{L}_{\mathrm{itm}}+\mathcal{L}_{\mathrm{icg}}+\mathcal{L}_{\mathrm{msp}}
\end{equation}

\subsection{Bootstraping Vision-to-Language Generative Learning via Semantic-aware Visual Object}

During the instruction fine-tuning stage, We connect MQ-Former with frozen image encoder and visual refiner to a LLM and apply a trainable projection matrix to convert the output query embedding $H_I$ into soft visual tokens, which maintain the same dimensional space as the word embeddings of the LLM. As the MQ-Former learns to integrate informative spatial and linguistic representations into the learned quries during pre-training, it can provide useful information to the LLM conducive to understanding and percepting the global visual features and semantic-aware visual objects. We employ low-rank adaption (LoRA)~\cite{hu2021lora} to adapt LLM by training multiple low-rank matrices for efficient alignment with human instruction and soft visual prompts. This facilitates the LLM in capturing multi-grained and multi-perspective visual information, which is conducive to the model integrating image details related to the instructions more precisely and mastering the capability to receive and output the spatial representation.

\begin{table}[!t]
\small
    \centering
    \fontsize{8}{11}\selectfont
    \caption{Details of Datasets used for pre-training and instruction fine-tuning. \vspace{-0.1in}}
\renewcommand{\arraystretch}{1.1}
\scalebox{1.0}{
    \begin{tabular}{p{1.4cm} p{6.2cm}}
         \toprule
        \textbf{Task} & \textbf{Dataset} \\
         \midrule
         \textit{Pre-training}  & LAION~\cite{laioncoco}, CC12M~\cite{changpinyo2021cc12m}, CC3M~\cite{sharma2018cc3m}, SBU~\cite{ordonez2011sbu},  \\
        \midrule
        \multicolumn{2}{l}{\textit{Instruction Tuning}} \\
        Captioning & COCO~\cite{chen2015coco}, CC3M~\cite{sharma2018cc3m}, MMC4~\cite{zhu2023mmc4}, AI Challenger \\
        Grounding & RefCOCO~\cite{kazemzadeh2014refcoco}, RefCOCO(+/g)~\cite{mao2016refcoco+/g}, GRIT~\cite{nguyen2022grit}, VG~\cite{krishna2017vg} \\
        General QA & VQAv2~\cite{goyal2017vqav2}, GQA~\cite{hudson2019gqa}, OK-VQA~\cite{marino2019okvqa}, DocVQA~\cite{mathew2021docvqa}\\
        Science QA & AI2D~\cite{hiippala2021ai2d}, SQA~\cite{lu2022sciqa}, TextVQA~\cite{sidorov2020textvqa} \\
        Chart QA & ChartQA~\cite{masry2022chartqa}, DVQA~\cite{kafle2018dvqa} \\
        Conv. & LLaVA~\cite{liu2023llava}, SVIT~\cite{zhao2023svit}, M3IT~\cite{li2023m3it}, LLaVAR~\cite{zhang2023llavar} MULTIINSTRUCT~\cite{xu2022multiinstruct},  Orca~\cite{mukherjee2023orca}, Alpace~\cite{alpaca} \\
    
         \bottomrule
    \end{tabular}}
    \label{tab:used_data}
    \vspace{-0.1in}
\end{table}

\section{Experiment Setting}

\begin{table*}[]
\centering
\fontsize{8}{11}\selectfont
\caption{Results on Image Captioning, General VQA and Text-oriented VQA datasets. Compared with prominent LVLMs, Lyrics achieves the best performance on 9/10 benchmarks. The best results are \textbf{bold} and the second-best results are \underline{underlined}. \vspace{-0.1in}}
\scalebox{1.0}{
\begin{tabular}{@{}l|ccc|ccccc|cc@{}}
\toprule
 \multirow{2}{*}{Model} & \multicolumn{3}{c|}{Image Captioning} & \multicolumn{5}{c}{General VQA} & \multicolumn{2}{c}{Text-oriented VQA}\\
   & \begin{tabular}[c]{@{}c@{}}COCO\end{tabular} & \begin{tabular}[c]{@{}c@{}}Nocaps\\ (0-shot)\end{tabular} & \begin{tabular}[c]{@{}c@{}}Flickr30K\\ (0-shot)\end{tabular} & VQAv2 & OKVQA & GQA & \begin{tabular}[c]{@{}c@{}}SciQA-Img\\ (0-shot)\end{tabular} & \begin{tabular}[c]{@{}c@{}}VizWiz\\ (0-shot)\end{tabular} & \begin{tabular}[c]{@{}c@{}}TextVQA\end{tabular} & \begin{tabular}[c]{@{}c@{}}OCR-VQA \\ (0-shot)\end{tabular} \\ \midrule
 Flamingo-9B &79.4 & - & 61.5 & 51.8 & 44.7 & - & - & 28.8 & 31.8 & - \\
 Flamingo-80B &84.3 & - & 67.2 & 56.3 & 50.6 & - & - & 31.6 & 35.0 & -  \\
 IDEFICS-9B (LLaMA-7B) & 46.0 & 36.8 & 27.3 & 50.9 & 38.4 & - & 44.2 & 35.5 & 25.9 & -  \\
 IDEFICS-80B (LLaMA-65B) & 91.8 & 65.0 & 53.7 & 60.0 & 45.2 & - & \underline{68.9} & 36.0 & 30.9 & -  \\
 BLIP-2 (Vicuna-13B) & - & 103.9 & 71.6 & 65.0 & 45.9 & 41.0 & 61.0 & 19.6 & 42.4 & - \\
 InstructBLIP (Vicuna-13B) & - & 121.9 & \underline{82.8} & - & - & 49.5 & 63.1 & 33.4 & 50.7 & -  \\
 Shikra (Vicuna-13B) & \underline{117.5}& - & 73.9 & 77.4 & 47.2 & - & - & -  & - & - \\
 Qwen-VL (Qwen-7B) & - & \underline{120.2} & 81.0 & \underline{78.2} & \underline{56.6} & \underline{57.5} & 68.2 & \textbf{38.9} & \underline{61.5} & \underline{70.5} \\  
  \midrule
\rowcolor[HTML]{F2F3F5} 
 \textbf{Lyrics (Vicuna-13B)} & \textbf{121.1} & \textbf{126.8} & \textbf{85.4} & \textbf{81.2} & \textbf{58.2} & \textbf{62.4} & \textbf{71.1} & \underline{37.6}  & \textbf{69.4} & \textbf{75.8}  \\
 \bottomrule
\end{tabular}
}
\label{tab:caption_vqa}
\end{table*}
 \begin{table*}[]
\centering
\fontsize{8}{11}\selectfont
\caption{Results on REC benchmarks. Generalist-VL models can directly generate the boundary boxes, while specialist models are specifically designed
for localization. Lyrics outperforms many generalist-VL models including OFA~\cite{wang2022ofa}, Shikra~\cite{chen2023shikra} and Qwen-VL~\cite{bai2023qwen}, and reduces the accuracy gap comparing to specialist models including UNINEXT~\cite{yan2023uninext} and G-DINO-L~\cite{zhang2022dino}. \vspace{-0.1in}}
\scalebox{1.2}{
\begin{tabular}{@{}l|l|ccccccccc@{}}
\toprule
\multirow{2}{*}{Model type} & \multirow{2}{*}{Model} & \multicolumn{3}{c}{RefCOCO} & \multicolumn{3}{c}{RefCOCO+} & \multicolumn{2}{c}{RefCOCOg} & \multirow{2}{*}{AVG}\\
 &  & val & test-A & test-B & val & test-A & test-B & val & test & \\  \midrule
\multirow{5}{*}{\begin{tabular}{l}Generalist \\ Models\end{tabular}} 
 & OFA-L* & 79.96 & 83.67 & 76.39 & 68.29 & 76.00 & 61.75 & 67.57 & 67.58 & 72.65 \\
 & Shikra (Vicuna-7B) & 87.01 & 90.61 & 80.24 & 81.60 & 87.36 & 72.12 & 82.27 & 82.19 &  82.93  \\
 & Shikra (Vicuna-13B) & 87.83 & 91.11 & 81.81 & 82.89 & 87.79 & 74.41 & 82.64 & 83.16 & 83.96 \\
 & Qwen-VL (Qwen-7B) & 88.55 & \textbf{92.27} & 84.51 & 82.82 & 88.59 & \textbf{76.79} & 85.96 & 86.32 & 85.73 \\ 
  & \textbf{Lyrics (Vicuna-13B)} & \textbf{90.69} & 92.08 & \textbf{86.03} & \textbf{82.89} & \textbf{89.77} & 76.72 & \textbf{87.23} & \textbf{88.26} & \textbf{86.71} \\ 
 \midrule
\multirow{2}{*}{\color{dt}\begin{tabular}{l}Specialist \\ Models\end{tabular}} & \color{dt}G-DINO-L & \color{dt}90.56 & \color{dt}93.19 & \color{dt}88.24 & \color{dt}82.75 & \color{dt}88.95 & \color{dt}75.92 & \color{dt}86.13 & \color{dt}87.02 & \color{dt} 86.60 \\
 & \color{dt}UNINEXT-H & \color{dt}92.64 & \color{dt}94.33 & \color{dt}91.46 & \color{dt}85.24 & \color{dt}89.63 & \color{dt}79.79 & \color{dt}88.73 & \color{dt}89.37 & \color{dt} 88.90 \\
 \bottomrule
\end{tabular}
}
\label{tab:grounding}
\vspace{-0.1in}
\end{table*}

\paragraph{Training Data}
In the pre-training stage, we use a large-scale, web-crawled set of image-text pairs and filter out low-relevant samples. In the instruction fine-tuning stage, we first use a wide range of publicly available vision-language datasets and transform them into instruction fine-tuning format for multi-task learning. Then, we introduce high-quality vision-language annotation and instruction-response data to enhance the instruction following and dialogue capabilities of Lyrics. We present the detailed description and statistics for each dataset in Table~\ref{tab:used_data}.

\paragraph{Implementation Detail}
For model settings, we choose ViT-L/14\\~\cite{dosovitskiy2020vit} initialized from pre-trained CLIP~\cite{radford2021clip} via contrastive learning as the image encoder. We build the visual refiner by combining Grounding-DINO-T~\cite{zhang2022dino} with 900 output object boxes and Swin-T backbone, SAM-HQ~\cite{kirillov2023sam} with MAE and pre-trained VIT-H image encoder, and RAM++~\cite{zhang2023ram} with Swin-B backbone. We use Vicuna-13B~\cite{vicuna2023}, an instruction-tuned variant from LLaMA~\cite{touvron2023llama}, as the foundation backbone. Throughout the entire training process, the image encoder and visual refiner remains frozen. We focus on training the MQ-Former and linear projection layer, and efficient fine-tuning the large language model using LoRA~\cite{hu2021lora}. With LoRA, we fine-tune the $\mathcal{W}_q$ and $\mathcal{W}_v$ via low-rank adaptation. We use images of size 224×224, augmented with random resized cropping and horizontal flipping. 

We use AdamW~\cite{loshchilov2018adamw} optimizer with $\beta_1=0.9, \beta_1=0.98$, and a weight decay of 0.05. We use a cosine learning rate scheduler to train our model decay with a peak learning rate of 1e-4 and a linear warmup ratio of 15\%. We train the Lyrics on 16xA100 GPUs for 800k steps in the vision-language representation alignment stage with a global batch size of 512, and 300k steps in the vision-to-language generative learning stage with a global batch size of 64.

\section{Experiment Result}
In this section, we conduct a comprehensive evaluation across various multi-modal tasks to thoroughly assess the visual understanding and generating capabilities of Lyrics, and compare our methods with the state-of-the-art visual-centric generalist models under zero-shot and few-shot settings, primarily including Flamingo~\cite{alayrac2022flamingo}, IDEFICS~\cite{laurencon2023obelics}, LLaVA~\cite{liu2023llava}, BLIP-2~\cite{li2023blip2}, InstructBLIP~\cite{dai2024instructblip}, Shikra~\cite{chen2023shikra}, Qwen-VL~\cite{bai2023qwen}, ShareGPT4V~\cite{chen2023sharegpt4v} and task-specific methods.

\subsection{Dataset and Evaluation Metrics}
We evaluate our model across a range of image captioning, VQA and REC benchmarks. For image captioning, we choose COCO~\cite{chen2015coco}, Nocaps~\cite{agrawal2019nocaps} and Flickr30K~\cite{plummer2015flickr30k} as benchmarks and report CIDEr score~\cite{vedantam2015cider} as metric. We consider five benchmarks including VQAv2\\~\cite{goyal2017vqav2}, OKVQA~\cite{marino2019okvqa}, GQA~\cite{hudson2019gqa}, ScienceQA (Image Set)~\cite{lu2022sciqa} and Vizwiz~\cite{gurari2018vizwiz} benchmarks for general VQA, two benchmarks including TextVQA~\cite{sidorov2020textvqa} and OCR-VQA~\cite{mishra2019ocrvqa} for text-oriented VQA, and evaluate the performance by matching the model’s response to the ground-truth and reporting top-1 accuracy. We use a sort of REC benchmarks such as RefCOCO~\cite{kazemzadeh2014refcoco}, RefCOCO+~\cite{mao2016refcoco+/g} and RefCOCOg~\cite{mao2016refcoco+/g} to verify the image understanding and localization capabilities. A predicted bounding box is considered as correct for reporting accuracy if its IOU between prediction and ground-truth is higher than 0.5. We use an open-ended approach with a greedy decoding strategy. We further conduct a comprehensive evaluation across 11 benchmark toolkits to thoroughly assess the multi-modal perception and conversation capabilities, which involve open-ended answers and factual assessments. Here we report the results in MathVista~\cite{lu2023mathvista}, MMMU~\cite{yue2023mmmu}, MME Perception (MME$^P$)~\cite{fu2023mme}, MME Cognition (MME$^C$)~\cite{fu2023mme}, MMBench (MMB)~\cite{liu2023mmbench}, MMBench-Chinese (MMB$^{CN}$)~\cite{liu2023mmbench}, SEED-Bench Image Part
(SEED$^I$)~\cite{li2023seed}, LLaVA-Bench In-the-Wild (LLaVA$^W$)~\cite{liu2023llava}, MM-Vet~\cite{yu2023mmvet}, QBench-Testset (QBench$^T$)~\cite{wu2023qbench} and HallusionBench (HallB)~\cite{guan2023hallusionbench}.

\subsection{Image Understanding Results}
\paragraph{\textbf{Image Captioning and Visual Question Answering.}}We first evaluate Lyrics on multiple image captioning and general VQA benchmarks. As demonstrated in Table~\ref{tab:caption_vqa}, We discover that Lyrics achieves the best performance across 7 out of 8 benchmarks, and demonstrate competitive results in the remaining VizWiz. Lyrics consistently surpasses its original backbone BLIP-2 by a significant margin across all benchmarks, and achieves competitive performance to Qwen-VL, which possesses a more robust LLM backbone and underwent more pre-training and instruction fine-tuning steps. For instance, we achieve the 121.1, 126.8 and 85.4 state-of-the-art CIDEr scores on three image captioning benchmarks, even outperforms previous generalist models with much more parameters (e.g., Flamingo and IDEFICS with 80B parameters). We achieve 62.1$\%$ average accuracy on all benchmarks for general VQA tasks, representing a relative improvement of 15.6$\%$ over BLIP-2. It indicates that the local visual features and spatial information provided by the visual refiner effectively facilitate fine-grained visual-language alignment, thus improving the model's ability to capture and respond to instruction-oriented visual objects. Furthermore, Table~\ref{tab:caption_vqa} also presents our experiment results on text-oriented VQA benchmarks, from which we can observe that Lyrics significantly outperforms the latest Qwen-VL by 7.9$\%$ and 5.3$\%$ on the TextVQA and OCR-VQA benchmarks. We believe that the improvement can be attributed to the introduction of semantic-aware visual objects extracted from MQ-Former, which facilitate the understanding of text within images.

\paragraph{\textbf{Referring Expression Comprehension.}} To demonstrate the fine-grained image comprehension and localization capabilities of our model, we examine the performance of various generalist models and specialist models on the REC task. As illustrated in Table~\ref{tab:grounding}, Lyrics achieves an average accuracy of 86.71$\%$ across 8 metrics on 3 benchmarks, surpassing the strong baseline Shikra~\cite{chen2023shikra} by 2.75$\%$ under the same LLM, and is on par with specialist model G-DINO-L. Compared to Shikra that directly employs spatial coordinates during the instruction fine-tuning stage to train the entire LLM (more than 13B trainable parameters), our improvement under the condition of lightweight training (merely 278M trainable paraeters) indicates that promoting semantic alignment between textualized spatial information and visual objects during the pre-training stage enables the promising performance in visual grounding.

\begin{table*}[t!]
\footnotesize
\centering
\fontsize{6.3}{11}\selectfont
\caption{Comparison with open-source SOTA methods on benchmark toolkits. Lyrics outperforms competitors in 9 out of 11 benchmarks and ranks second in the others. The best results are \textbf{bold} and the second-best results are \underline{underlined}. \vspace{-0.1in}} 
\scalebox{1.28}{
\setlength{\tabcolsep}{1.0mm}{
\begin{tabular}{l|ccccccccccc}
\toprule
 
Method &  MathVista & MMMU & MME$^{P}$ & MME$^{C}$ & MMB & MMB$^{CN}$ & SEED$^{I}$ & LLaVA$^{W}$ & QBench$^{T}$ & MM-Vet & HallB  \\ 
\midrule
        BLIP-2 (FLAN-T5) & - & 35.7 & 1293.8 & 290.0 & - & - & 46.4 & 38.1 & - & 22.4 & -  \\ 
        InstructBLIP (Vicuna-7B) & 25.3 & 30.6 & - & - & 36.0 & 23.7 & 53.4 & 60.9 & 55.9 & 26.2 & 53.6  \\ 
        IDEFICS-80B (LLaMA-65B) & 26.2 & 24.0 & - & - & 54.5 & 38.1 & 52.0 & 56.9 & - & 39.7 & 46.1  \\ 
        Qwen-VL-Chat (Qwen-7B) & \underline{33.8} & 35.9 & 1487.5 & 360.7 & 60.6 & 56.7 & 58.2 & 67.7 & \underline{61.7} & \textbf{47.3} & \underline{56.4}  \\ 
        LLaVA (Vicuna-7B) & 23.7 & 32.3 & 807.0 & 247.9 & 34.1 & 14.1 & 25.5 & 63.0 & 54.7 & 26.7 & 44.1  \\ 
        LLaVA-1.5 (Vicuna-13B) & 26.1 & 36.4 & 1531.3 & 295.4 & 67.7 & \textbf{63.6} & 68.2 & 70.7 & 61.4 & 35.4 & 46.7  \\ 
        ShareGPT4V (Vicuna-7B) & 25.8 & \underline{36.6} & \underline{1567.4} & \underline{376.4} & \underline{68.8} & 62.2 & \underline{69.7} & \underline{72.6} & - & 37.6 & 49.8  \\ 

\midrule
\rowcolor[HTML]{F2F3F5} 
        \textbf{Lyrics (Vicuna-13B)} & \textbf{39.4} & \textbf{40.2} & \textbf{1597.3} & \textbf{431.6} &\textbf{75.3} & \underline{62.4} & \textbf{71.8} & \textbf{76.9} & \textbf{72.5} & \underline{46.3} & \textbf{62.6} \\

 \bottomrule
\end{tabular} }}
\vspace{-0.1in}
\label{tab:entire_comp} 
\end{table*}

\subsection{Multi-Modal Benchmark Toolkit Results}
In Table~\ref{tab:entire_comp}, we present a quantitative comparison between our proposed Lyrics model with existing SOTA LVLMs. Specifically, Lyrics outperforms the previously best-performing ShareGPT4V model by 3.6, 6.5 and 4.3 points on the MMMU, MMB, LLaVA$^W$ benchmarks, demonstrating superior capabilities in tasks such as detailed description and complex reasoning. On the MME Benchmark, Lyrics achieves the highest scores in both perception (P) and cognition (C) capabilities, surpassing Qwen-VL-Chat in by 89.8 and 70.9 points, which was trained on 1.4 billion data. In the low-level image assessment QBench and multi-level image assessment with 14K questions SEED benchmarks, Lyrics achieves the highest score of 72.5\% and 71.8\%, 10.8\% and 2.1\% higher than the second-ranked LVLMs. which can be attributed to the diversity of our constructed dataset. Notably, Lyrics achieves a significant improvements on the MathVista and HallB benchmarks, demonstrating that the visual objects provided by the visual refiner can enhance the model's capability to perceive real symbols and eliminate visual hallucinations.

\begin{figure}[t]
    \centering
    \includegraphics[width=1\linewidth, trim=0 10 0 0]
    {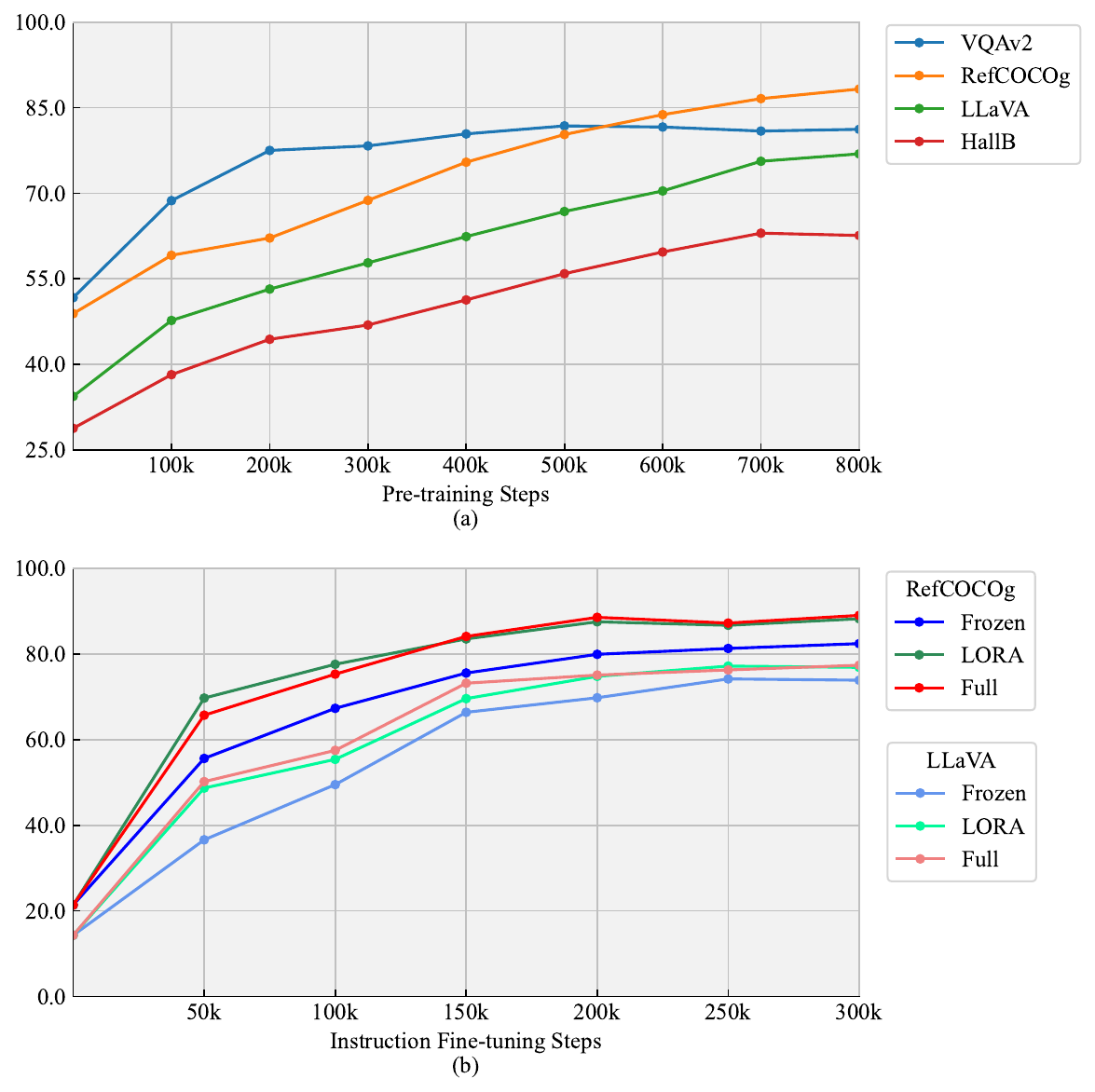}
    \vspace{-2em}
    \caption{\textbf{(a)} The pre-training data scaling 
    performance on VQAv2, RefCOCOg (testset), LLaVA-Bench and HallusionBench. \textbf{(b)} The comparison of full, LoRA and frozen training in instruction fine-tuning stage.}
    \label{fig:Few_shot_learning_results}
    \vspace{-0.05in}
\end{figure}

\begin{table}[!t]
    \centering
    \fontsize{8}{11}\selectfont
\caption{Ablation study on model architecture. \vspace{-0.1in}}
\scalebox{1.06}{
    \begin{tabular}{l|cccc}
         \toprule
        Architecture & VQAv2 & RefCOCOg$T$ & LLaVA$^W$ & HallB \\
         \midrule
        \rowcolor[HTML]{F2F3F5}
         \textbf{Lyrics}   &  \textbf{81.2}  &  \textbf{88.26}  &  \textbf{76.9}  &  \textbf{62.6} \\
         w/o ViT  &  78.5  &  86.11  &  74.6  &  60.1 \\
         w/o ODM  &  80.6  &  87.55  &  76.1  &  60.7 \\
         w/o SSM  &  80.8  &  87.08  &  76.5  &  59.2 \\
         w/o VR   &  77.2  &  83.25  &  73.3  &  56.8 \\  
         \bottomrule
    \end{tabular}}
    \label{tab:ablation_model_architecture}
    \vspace{-0.1in}
\end{table}

\subsection{Ablation Study}
In this section, we conduct ablation studies on the model architecture and training strategies to investigate the impact of semantic-aware visual objects and fine-grained representation alignment on the performance of Lyrics.

\paragraph{\textbf{Model Architecture.}} As illustrated in Table~\ref{tab:ablation_model_architecture}, we investigate the performance degradation of Lyrics on four benchmarks following the removal of the visual encoder (w/o ViT), object detection module (w/o ODM), semantic segmentation module (w/o SSM) and visual refiner (w/o VR). Generally, we use blank images to replace the original images as input for the specified module to represent the removal of the module. Firstly, the elimination of the ViT leads to performance declines across all tasks, attributable to the absence of global visual feature. Furthermore, We can observe that relying solely on either object detection or semantic segmentation modules results in insufficient local visual information. However, the concurrent removal of both leads to significant performance degradation across all datasets. Particularly, there is a 5.01\% and 5.80\% decrease in performance on the RefCOCOg and HallB benchmarks, respectively, demonstrating the significance of directly learning visual objects to grasp regional information.

\begin{figure*}[t]
    \centering
    \includegraphics[width=1\linewidth]{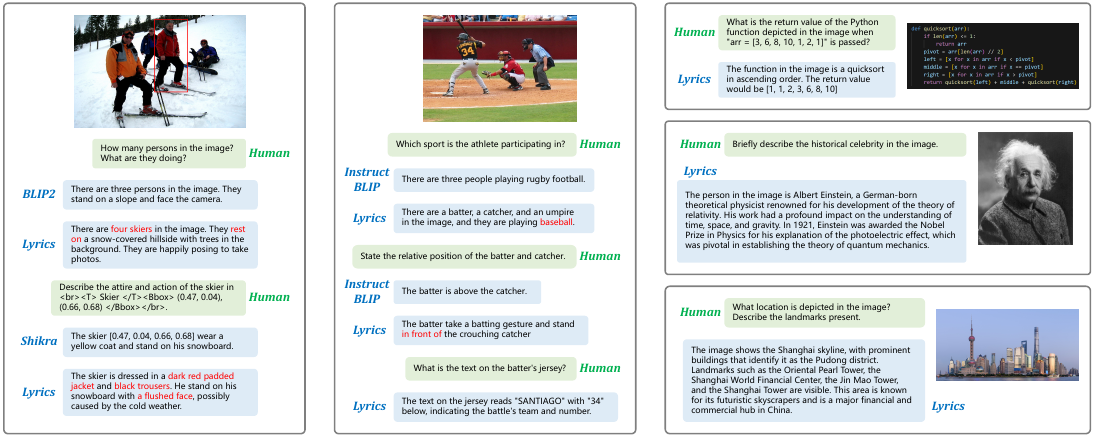}
    \caption{Examples for multi-modal capabilities of Lyrics, We showcase that our method is capable of various visual-centric tasks, including multi-turn visual conversation, visual scene understanding and reasoning, commonsense-grounded image description, referential dialogue.}
    \label{fig:case}
        \vspace{-0.1in}
\end{figure*}

\paragraph{\textbf{Training Strategy. }} In Figure (a), we present our investigation into the required quantity of high-quality image captions in the pre-training stage. Without vision-language alignment, Lyrics suffers from catastrophic forgetting where performance drastically degrades following instruction fine-tuning, and the model shows consistent gain with more pre-training data. Concurrently, with the increase in pre-training steps, the model can quickly adapt to straightforward VQA tasks (i.e., VQAv2) in instruction fine-tuning stage, whereas for complex visual reasoning and referential dialogue tasks (i.e., RefCOCOg$^T$, LLaVA$^W$ and HallB), effective fitting requires sufficient visual-language alignment beforehand. In Figure (b), we conduct a comprehensive comparison of training the LLM, visual encoder and visual refiner with frozen and full parameters during the instruction fine-tuning stage, as well as partially training LLM with LoRA strategy. The training results demonstrate that freezing all parameters diminishes the fitting speed and performance, primarily due to the lack of self-adjustment limiting the LLM to receive visual signals. Furthermore, the fitting trends between using LoRA and full-parameter training are remarkably similar, confirming that adequate visual-language alignment ensures lightweight instruction fine-tuning is sufficient for the LVLM to master various vision-to-language dialogue scenarios.

\subsection{Qualitative Results}
\label{sec:case}
We further provide the qualitative results for a complementary understanding of the instructed zero-shot image-to-text generation capability of Lyrics. As illustrated in Figure~\ref{fig:case}, we present the responses of Lyrics and various mainstream LVLMs, such as BLIP2~\cite{li2023blip2}, InstructBLIP~\cite{dai2024instructblip} and Shikra~\cite{chen2023shikra}, under the same instruction and image inputs. In the absence of fine-grained visual signals that necessary for performing counting, discerning colors, recognizing actions and judging position, previous methods fail to accurately capture detailed information of visual objects involved in the instructions. We observe that with the visual refiner, Lyrics can effectively avoid visual hallucinations and factual errors. It is reasonable to infer that Lyrics can understand and perceive the visual objects within the image via two-stage fine-grained vision-language representation alignment and generative learning. For example, Lyrics can leverage spatial information and local visual feature provided by the visual refiner to perceive the number, color and motion of visual objects contained in images, as well as to explore the relative positions between the perceived visual objects. Therefore, in the first case, Lyrics can identify that \textit{there are four skiers} in the image who are in a \textit{resting state}, and indicate \textit{a flushed face} of the skier and \textit{he wear dark red padded jacket and black trouser} within specific spatial coordinates via referential dialogue. Moreover, Lyrics effectively inherit the commonsense understanding and logical reasoning capabilities of LLMs, which enable the model to deduce the symbolic meaning of text and the result of code execution. We also discover that our method impressively identifies objective entities, such as notable figures and locations, indicating that the knowledge of LLMs is effectively fed back to MQ-Former during the process of instruction fine-tuning. More examples are displayed in supplementary materials.

 \begin{table}[]
\centering
\fontsize{8}{11}\selectfont
\caption{In-context few-shot learning results of prominent LVLMs on VQAv2, RefCOCOg (testset) and LLaVA-Bench. \vspace{-0.1in}}
\scalebox{1.05}{
\begin{tabular}{@{}l|l|cccc@{}}
\toprule
\multicolumn{2}{c|}{\multirow{2}{*}{Dataset}} & \multicolumn{3}{c}{Model} \\
\multicolumn{1}{c}{}& & BLIP2 & Shikra & Qwen-VL & Lyrics\\  \midrule
\multirow{3}{*}{VQAv2}  & 0-shot & 65.0 & 77.4 & 78.2 & 81.2 \\ 
                        & 1-shot & 67.4 & 79.7 & 80.5 & 85.2 \\
                        & 4-shot & 72.5 & 81.2 & 81.3 & \textbf{86.1} \\
 \midrule
 \multirow{3}{*}{RefCOCOg$^T$}  & 0-shot & - & 83.16 & 86.32 & 88.26 \\ 
                                & 1-shot & - & 84.29 & 87.13 & \textbf{91.39} \\
                                & 4-shot & - & 85.02 & 87.88 & 91.36 \\
 \midrule
 \multirow{3}{*}{LLaVA$^W$} & 0-shot & 38.1 & - & 67.7 & 76.9 \\ 
                            & 1-shot & 40.6 & - & 68.2 & \textbf{79.1} \\
                            & 4-shot & 43.3 & - & 68.6 & 78.6 \\
 \bottomrule
\end{tabular}
}

\label{tab:fewshot_learning}
\vspace{-0.2in}
\end{table}

\subsection{Few-shot Learning on VL Tasks}

To further verify the efficient learning and knowledge generalization of Lyrics, we conduct in-context few-shot learning on the VQAv2~\cite{goyal2017vqav2}, RefCOCOg (testset)~\cite{marino2019okvqa} and LLaVA-Bench~\cite{liu2023llava} datasets. Note that we adopt naive random sample to construct the few-shot exemplars, and report the averaged scores for five different seeds. As illustrated in Table~\ref{tab:fewshot_learning}, with a similar number of parameters, Lyrics exhibits stable performance peak and upward trend across various visual-language tasks, even when compared to LVLMs with powerful backbones. Notably, with one in-context learning sample, Lyrics achieves significant performance improvements across various tasks. Compared to existing methods, our model more effectively utilizes knowledge of instruction following within a short context window, demonstrating that diverse visual feature priors contribute to promoting autonomous segmentation of task frameworks by the model.

\section{Conclusion}
In this paper, we propose Lyrics, a two-stage fine-grained pre-training and instruction fine-tuning framework towards the generalist LVLM. We introduce a visual refiner designed to extract abstract local visual feature and concrete spatial information, which is comprised of an image tagging module, an object detection module and a semantic segmentation module. We first connect the Multi-scale Querying Transformer (MQ-Former) to frozen image encoder and visual refiner and bootstrap vision-language representation alignment via multi-task pre-training. Then, we connect the MQ-Former to the LLMs to bootstrap vision-to-language generative learning via semantic-aware object. Lyrics achieves impressive results across various vision-language tasks, and demonstrates a real-world dialogue capability in commonsense-grounded image description, visual scene understanding and reasoning, referential dialogue.

{
    \small
    \bibliographystyle{ieeenat_fullname}
    \bibliography{main}
}


\clearpage
\appendix

\section{Instruction Templates for Instruction Fine-tuning and Zero-shot Inference}
\label{appendix:zeroshot_instruction}
Image captioning and visual question answering (VQA) are conventional tasks for vision-language models. Specifically, \textbf{Image Captioning} aims to generate a descriptive text (caption) to describe the given image, while \textbf{Grounded Captioning} aims to generate a descriptive text (caption) to describe the specified regions of an image. \textbf{General Visual Question Answering} requires models to understand the content of image and question to generate answer, while \textbf{Text-oriented Visual Question Answering} aim at reading and understanding scene text within images for question answering. In calculating metrics in the paper, we regard the \textbf{Referring expression comprehension} (REC) as a \textbf{Visual Grounding} task that locates specific objects referred to by natural language expressions. The expression provides high-level concepts of relevant visual and contextual patterns. Additionally, we formulate instruction and response formats for two special derivative tasks. Firstly, \textbf{Referential Dialogue} focuses on conducting image captioning or visual question answering tasks targeting the specific objects within the image, which expects the model to mention the coordinates and tags of the relevant objects in both the instruction and response. Secondly, \textbf{Multiple-choice Visual Question Answering} task provides several candidate choices for a question within the instruction and require the model to select one of them as the response. We separate options with the alphabetical order, e.g. (a) blue (b) yellow (c) pink (d) black. Around these different categories of multi-modal tasks, drawing inspiration from InstructBLIP~\cite{dai2024instructblip} and Shikra~\cite{chen2023shikra}, we formulate various instructions for each task. For the pure-text auto-regression and multi-modal instruction tasks, we directly utilize the formats originally inherent in the dataset. As illustrated in Table~\ref{tab:appendix_instruct_template}, We provide instructions used for instruction fine-tuning and zero-shot inference.

\begin{table*}[h!]
\centering
\scalebox{0.96}{%
\begin{tabular}{l|l}
\toprule
Task & Instruction Template \\ \midrule
\begin{tabular}[c]{@{}l@{}}Image\\ Captioning\end{tabular} & \begin{tabular}[c]{@{}l@{}} \textless{}Image\textgreater Write a short description for the image.\\ \textless{}Image\textgreater Write a description for the image.\\ \textless{}Image\textgreater Provide a description of what is presented in the photo.\\ \textless{}Image\textgreater Briefly describe the content of the image.\\ \textless{}Image\textgreater Look at the image and describe what you see in a simple and clear manner. \\ \textless{}Image\textgreater Could you use a few words to describe what you perceive in the photo?\\ \textless{}Image\textgreater Please provide a short depiction of the picture.\\ \textless{}Image\textgreater Summarize what this image depicts in a simple and concise manner.\\ \textless{}Image\textgreater Provide a simple and clear description of the image, suitable for all audiences. \end{tabular} \\ \midrule

\begin{tabular}[c]{@{}l@{}}Visual\\ Question \\ Answering \end{tabular} & \begin{tabular}[c]{@{}l@{}}\textless{}Image\textgreater \{Question\}\\ \textless{}Image\textgreater Question: \{Question\}\\ \textless{}Image\textgreater Question: \{Question\} Answer:\\ \textless{}Image\textgreater Given the image, answer the following question: \{Question\} \\ \textless{}Image\textgreater With the aid of the following image, offer a straightforward, short response to: \{Question\}.\\ \textless{}Image\textgreater Based on the image, respond to this question with a short answer: \{Question\}. Answer:\\ \textless{}Image\textgreater Use the provided image to answer the question as short as possible: \{Question\}\\ \textless{}Image\textgreater What is the answer to the following question? \{Question\}\\ \textless{}Image\textgreater Refer to the information in the image to provide a minimalist answer to: \{Question\} \end{tabular} \\ \midrule

\begin{tabular}[c]{@{}l@{}} Text-oriented \\ Visual\\ Question \\ Answering \end{tabular} & \begin{tabular}[c]{@{}l@{}}\textless{}Image\textgreater Question: \{Question\}\\ \textless{}Image\textgreater Question: \{Question\} Answer:\\ \textless{}Image\textgreater Analyze the textual content in this image and provide a short answer to: \{Question\}. \\ \textless{}Image\textgreater Look at the text in the image provided and succinctly answer: \{Question\}.\\ \textless{}Image\textgreater With the help of text in the following image, offer a simple, short response to: \{Question\}.\\ \textless{}Image\textgreater Refer to the textual data in the image to provide a brief answer to: \{Question\}. \\ \end{tabular} \\ \midrule

\begin{tabular}[c]{@{}l@{}} Grounded \\ Captioning \end{tabular} & \begin{tabular}[c]{@{}l@{}}\textless{}Image\textgreater Write a description for the target object in the image. \\ \textless{}Image\textgreater Provide a short caption focusing on the highlighted object in this image. \\ \textless{}Image\textgreater Describe the specific object indicated in the following image, keeping the description brief. \\ \textless{}Image\textgreater Explain what the object marked in the image is, using a concise description. \\ \textless{}Image\textgreater Identify and describe the key object in this image, using a short and clear description. \\ \end{tabular} \\ \midrule

\begin{tabular}[c]{@{}l@{}} Referring \\ Expression \\ Comprehension \end{tabular} & \begin{tabular}[c]{@{}l@{}}\textless{}Image\textgreater In the given image, could you find and tell me the coordinates of \{$Tag$\}? \\ \textless{}Image\textgreater In the coordinate \{$Bbox$\} of the image, can you observe the object \{$Tag$\}. \\ \textless{}Image\textgreater Locate the \{$Tag$\} in this image and provide a brief description of its position. \\ \textless{}Image\textgreater Confirm the presence of \{$Tag$\} in the bounding box \{$Bbox$\} in the image. \\
\textless{}Image\textgreater Search for \{$Tag$\} in the image and give its coordinates if found. \\
\textless{}Image\textgreater Can you find the spatial location or coordinates of \{$Tag$\} in the image shown here? 
 \\ \end{tabular}
\\ \midrule

\begin{tabular}[c]{@{}l@{}} Referential \\ Dialogue \end{tabular} & \begin{tabular}[c]{@{}l@{}}\textless{}Image\textgreater Focus on the object \{$Tag\ \& \ Bbox$\} in the image, and answer the question: \{Question\}. \\ 
\textless{}Image\textgreater Could you provide a descriptive caption for the object \{$Tag\ \& \ Bbox$\} in the image? \\ \textless{}Image\textgreater Regarding the object specified as \{$Tag\ \& \ Bbox$\}, please respond to: \{Question\}. \\ \textless{}Image\textgreater Explain the features or details of the object identified by \{$Tag\ \& \ Bbox$\} in the image. \\
\textless{}Image\textgreater Create a caption that describes the area or object marked as \{$Tag\ \& \ Bbox$\} in the image. \\
\textless{}Image\textgreater Refer to the object \{$Tag\ \& \ Bbox$\} in the image, and provide an answer to: \{Question\}. \\
\end{tabular}
\\ \midrule

\begin{tabular}[c]{@{}l@{}} Multi-choice \\ Visual\\ Question \\ Answering  \end{tabular} & \begin{tabular}[c]{@{}l@{}}\textless{}Image\textgreater Question: \{Question\} Options: \{Option\}. Answer: \\ 
\textless{}Image\textgreater For the question: \{Question\}, choose the most suitable answer from options: \{Option\}. \\ \textless{}Image\textgreater Examine the image and answer the question: \{Question\}. Your choices are: \{Option\}. \\ \textless{}Image\textgreater Respond to the question: \{Question\} among options: \{Option\}, select your response: \\
\textless{}Image\textgreater Consider the question: \{Question\} and options: \{Option\}. Please provide your answer: \\
\end{tabular}
\\

\bottomrule
\end{tabular}%
}
\vspace{2pt}
\caption{Instruction templates used for transforming the conventional vision-language datasets into instruction tuning data.}
\label{tab:appendix_instruct_template}
\end{table*}

\begin{figure*}[b]
    \centering
    \includegraphics[width=1\linewidth, trim=0 10 0 0]{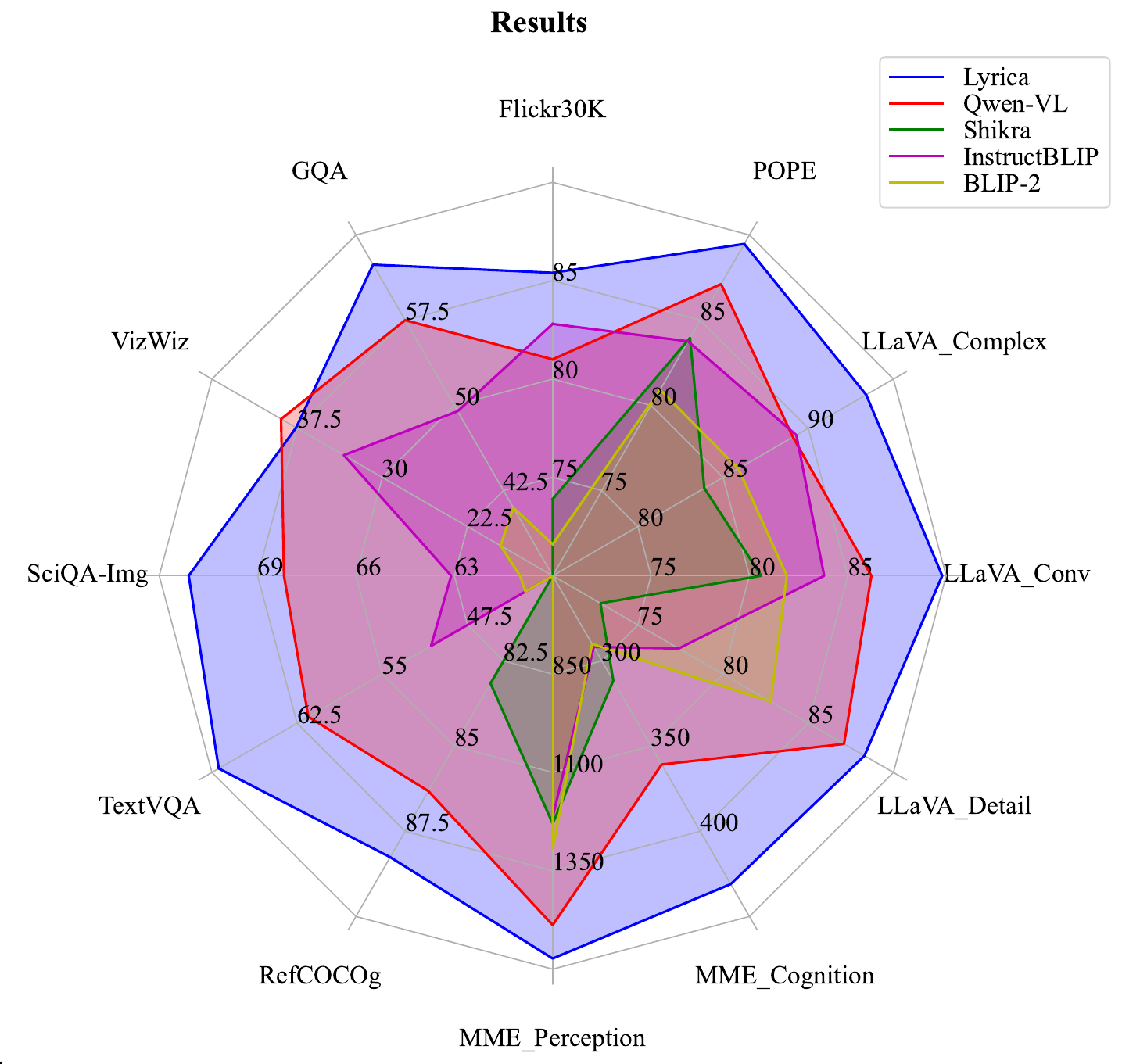}
    \caption{Our Lyrics achieves state-of-the-art performance on a broad range of vision-language tasks compared with other generalist models.}
    \label{fig:graph_result}
    \vspace{-0.2in}
\end{figure*}

\end{document}


\title{Lyrics: Boosting Fine-grained Language-Vision Alignment via \\ Semantic-aware Visual Objects}

\setcounter{page}{1}
\maketitlesupplementary
\appendix

\section{Dataset Details}
\label{appendix_dataset_detail}

In the first stage of pre-training, We primarily utilize a large-scale, weakly-labeled, web-crawled image-text pair dataset, which is composed of several publicly accessible data sources. As summarized in Table~\ref{tab:pretraining_data}, We filter out image-text pairs with a similarity below 25$\%$ through an internally trained CLIP~\cite{radford2021clip} model and use the cleaned data for pre-training. We clean these noisy data by several steps:

During instruction fine-tuning stage, we introduce high-quality and fine-grained vision-language annotation data. As summarized in Table~\ref{tab:multitask_data}, we divide the collected datasets into 7 tasks and only take the train dataset to fine-tune Lyrics simultaneously. We transform 24 datasets into the instruction tuning format and group them into 6 task categories. For text generation, we use the collected corpus Orca~\cite{mukherjee2023orca} and Alpaca~\cite{alpaca} to recover the language generation ability. Image captioning data is COCO~\cite{chen2015coco}, cleaned CC3M and SBU. We use a mixture of publicly available data for the VQA task which includes GQA~\cite{hudson2019gqa}, VQAv2~\cite{goyal2017vqav2}, OCR-VQA~\cite{mishra2019ocrvqa}, OK-VQA~\cite{marino2019okvqa}, AOK-VQA~\cite{schwenk2022aokvqa}, DocVQA~\cite{mathew2021docvqa}, TextVQA~\cite{sidorov2020textvqa} and ChartQA~\cite{masry2022chartqa}. For the referring expression comprehension and grounded captioning duality tasks, we uniformly construct training samples from GRIT~\cite{nguyen2022grit}, Visual Genome~\cite{krishna2017vg}, RefCOCO~\cite{kazemzadeh2014refcoco}, RefCOCO+~\cite{mao2016refcoco+/g} and RefCOCOg~\cite{mao2016refcoco+/g}. In order to improve the instruction following and dialogue capabilities, we further collect various instruction fine-tuning datasets for real-world scenarios, including LLaVA~\cite{liu2023llava}, SVIT~\cite{zhao2023svit}, M3IT~\cite{li2023m3it}, MULTIINSTRUCT~\cite{xu2022multiinstruct} and LLaVAR~\cite{zhang2023llavar}.

\section{Summary of the Evaluation Benchmarks and Overfitting Authentication}
We provide a detailed summary of the used evaluation benchmarks and corresponding metrics in Table~\ref{tab:benchmark}. Since we aim to stimulate the model's comprehension and generation capabilities in primary vision-language tasks, we believe that incorporating the training set of a specific dataset into the extensive training process will not result in the model overfitting to that dataset's performance patterns. Consequently, we define evaluations that incorporate the training set into the instruction fine-tuning stage as few-shot evaluations, and those that do not include the training set in the instruction fine-tuning phase as zero-shot evaluations.

Therefore, we conduct an ablation study on the Nocaps~\cite{agrawal2019nocaps}, Flickr30k~\cite{plummer2015flickr30k}, VQAv2~\cite{goyal2017vqav2}, OKVQA~\cite{marino2019okvqa}, GQA~\cite{hudson2019gqa}, TextVQA~\cite{sidorov2020textvqa} and RefCOCOg~\cite{mao2016refcoco+/g} datasets, in which we remove the train split of the corresponding dataset from the instruction fine-tuning stage, aiming to achieve zero-shot evaluation. As illustrated in Table~\ref{tab:ablation_appendix}, removing the train split of a specific dataset during the instruction fine-tuning stage does not significantly impact the model's performance on that dataset's test split. For example, by excluding the train split of the Visual Question Answering dataset VQAv2 from our vision-language generative learning, the model train in this manner exhibits only a 1.3\% lower performance on the zero-shot evaluation of VQAv2 compared to the few-shot evaluation. Interestingly, this approach even yields slight performance improvements on some datasets. This phenomenon demonstrates that with adequate training on diverse datasets, LVLMs can develop robustness and generalization, preventing them from being overly specialized and overfitting to certain specific tasks and formats.

\begin{table}[!t]
    \centering
\scalebox{1.0}{
    \begin{tabular}{l ccc}
         \toprule
        \textbf{Dataset} & \textbf{Original} & \textbf{Cleaned} & \textbf{Remaining$\%$} \\
         \midrule
         LAION-en     & 2B       & 200M & 10$\%$ \\
         LAION-COCO   & 600M     & 300M & 50$\%$ \\
         CC12M        & 12M      & 6M   & 50$\%$ \\
         CC3M         & 3M       & 3M   & 100$\%$ \\
         SBU          & 1M       & 0.9M & 90$\%$ \\
         COCO Caption & 0.6M     & 0.6M & 100$\%$ \\
         \bottomrule
    \end{tabular}}
        \caption{Details of web-crawl datasets in vision-language representation alignment stage. LAION-COCO~\citep{laioncoco} is a synthetic dataset generated from LAION-en~\cite{laion5b}. CC12M~\citep{changpinyo2021cc12m}, CC3M~\citep{sharma2018cc3m}, SBU~\citep{ordonez2011sbu} and COCO Caption~\citep{chen2015coco} are academic caption datasets.}
    \label{tab:pretraining_data}
\end{table}

\section{Instruction Templates for Instruction Fine-tuning and Zero-shot Inference}
\label{appendix:zeroshot_instruction}
Image captioning and visual question answering (VQA) are conventional tasks for vision-language models. Specifically, \textbf{Image Captioning} aims to generate a descriptive text (caption) to describe the given image, while \textbf{Grounded Captioning} aims to generate a descriptive text (caption) to describe the specified regions of an image. \textbf{General Visual Question Answering} requires models to understand the content of image and question to generate answer, while \textbf{Text-oriented Visual Question Answering} aim at reading and understanding scene text within images for question answering. In calculating metrics in the paper, we regard the \textbf{Referring expression comprehension} (REC) as a \textbf{Visual Grounding} task that locates specific objects referred to by natural language expressions. The expression provides high-level concepts of relevant visual and contextual patterns. Additionally, we formulate instruction and response formats for two special derivative tasks. Firstly, \textbf{Referential Dialogue} focuses on conducting image captioning or visual question answering tasks targeting the specific objects within the image, which expects the model to mention the coordinates and tags of the relevant objects in both the instruction and response. Secondly, \textbf{Multiple-choice Visual Question Answering} task provides several candidate choices for a question within the instruction and require the model to select one of them as the response. We separate options with the alphabetical order, e.g. (a) blue (b) yellow (c) pink (d) black. Around these different categories of multi-modal tasks, drawing inspiration from InstructBLIP~\cite{dai2023instructblip} and Shikra~\cite{chen2023shikra}, we formulate various instructions for each task. For the pure-text auto-regression and multi-modal instruction tasks, we directly utilize the formats originally inherent in the dataset. As illustrated in Table~\ref{tab:appendix_instruct_template}, We provide instructions used for instruction fine-tuning and zero-shot inference.

\section{Additional Experiment Details on Multi-modal Leaderboards}
\label{appendix:leaderboard}
To have a systematic understanding of the performance of Lyrics, we
aim to leverage a quantitative metric in measuring the model’s instruction-following capability under the real-world conversation scenarios. Therefore, 2e evaluate the performance of Lyrics on three leaderboards: \textbf{LLaVA}~\cite{liu2023llava}, \textbf{MME}~\cite{fu2023mme}, and \textbf{POPE}, and compare it with mainstream generalist models such as BLIP2~\cite{li2023blip2}, InstructBLIP~\cite{dai2023instructblip}, Shikra~\cite{chen2023shikra}, and Qwen-VL~\cite{bai2023qwen}. 

The LLaVA leaderboard utilizes GPT-4~\cite{openai2023gpt4} to assess the quality of the responses generated by the LVLMs. Specifically, we randomly select 30 images from the COCO validation split, and generate three types of question (conversation, detailed description, complex reasoning) using the proposed data generation pipeline. LVLMs predicts the answers based on the question and input image. GPT-4 makes a reference prediction based on the question, ground-truth bounding boxes and captions, marking an upper bound of the teacher model. After obtaining the response from both models, we feed the question, visual information (in the format of captions and bounding boxes), and the generated responses from both assistants, to the GPT-4. GPT-4 evaluates the helpfulness, relevance, accuracy, and level of details of the responses from the assistants, and give an overall score on a scale of 1 to 10, where a higher score indicates better overall performance. GPT-4 is also asked to provide a comprehensive explanation the evaluation, for us to better understand the models. 

The MME leaderboard covers the examination of perception and cognition abilities across 14 subtasks. The perception includes the recognition of coarse-grained and fine-grained objects. The former
identifies the existence, count, position, and color of objects. The latter recognizes movie posters, celebrities, scenes, landmarks, and artworks. The cognition includes commonsense reasoning, numerical calculation, text translation, and code reasoning. Since this leaderboard limits the output of model to two types (“yes” or “no”), it measure the metrics of accuracy and accuracy+ to validate the performance. The former is calculated based on each question, while the latter is based on each image where both of the two questions need to be answered correctly. We calculate the score of a subtask based on the sum of accuracy and accuracy+.

The POPE leaderboard formulates the evaluation of object hallucination as a binary classification task that prompts LVLMs to output “Yes” or “No”, e.g., “Is there a chair in the image?”. Questions whose answers are “Yes” can be directly built using ground-truth objects, while questions with the answer “No” can be built by sampling from negative objects. It employs three sampling strategies to verify whether LVLMs are prone to hallucinate specific objects, including: Random Sampling, which randomly samples the objects that do not exist in the image; Popular Sampling, which selects the top-k most frequent objects in the whole image dastaset that do not exist in the current image; Adversarial Sampling, which first ranks all objects according to their co-occurring frequencies with the ground-truth objects and then selects the top-k frequent ones that do not exist in the image.

As illustrated in Table~\ref{tab:real-world metrics}, in various categories and objectives of visual-language instruction tasks, Lyrics surpasses other LVLMs by a large margin, from which we can draw the following conclusions: (1) In the LLaVA leaderboard, Lyrics demonstrates scores approaching those of GPT-4. This indicates that the model, in addition to inheriting the logical reasoning and language expression capabilities of LLMs, has successfully integrate substantial visual signals into its understanding. (2) The significant performance improvement of Lyrics on MME and POPE leaderboards indicates its ability to largely mitigate the effects of visual illusions. It is likely attributed to the semantic-aware visual objects introduced by the Visual Refiner, which enables the model to have a clearer understanding of concepts related to quantity, color, position, and morphology in the images. (3) Lyrics demonstrates robust instruction-following capability by adapting to variations in instructions, enabling it to execute visual-language tasks in various scenarios, such as visual-centric description, visual scene understanding and reasoning, commonsense-grounded image perception, referential dialogue.

\begin{table*}[!b]
    \centering
\scalebox{0.95}{
    \begin{tabular}{l c}
         \toprule
         \textbf{Task} &  \textbf{Dataset} \\
         \midrule
         Image Captioning   & COCO 2014, COCO 2017, CC3M, SBU \\
         Grounded Captioning &  GRIT, Visual Genome, RefCOCO, RefCOCO+, RefCOCOg \\
         Visual Question Answering & GQA, VQAv2, OCR-VQA, OK-VQA, AOK-VQA, DocVQA, TextVQA, ChartQA \\
         Referring Expression Comprehension & GRIT, Visual Genome, RefCOCO, RefCOCO+, RefCOCOg \\
         Multi-modal Instruction & LLaVA, SVIT, M3IT, MULTIINSTRUCT, LLaVAR \\
         Pure-text Auto-regression & Orca, Alpaca \\
         \bottomrule
    \end{tabular}
    }
        \caption{Details of instruction-tuning data in vision-language generative learning stage.
    }
    \label{tab:multitask_data}
\vspace{1em}  
\end{table*}

\begin{table*}[!h]
    \centering
    \scriptsize
    \scalebox{1.16}{
    \begin{tabular}{l|l|l|l|l}
        \toprule
         Task & Dataset & Description & Split & Metric  \\
         \midrule
         \multirow{2}{*}{\makecell[l]{Image \\ Captioning}} & Nocaps & Captioning of natural images & val & CIDEr($\uparrow$) \\
         & Flickr30K & Captioning of natural images & karpathy-test & CIDEr($\uparrow$) \\
         \midrule
         \multirow{5}{*}{\makecell[l]{General \\ VQA}} & VQAv2 & VQA on natural images & test-dev & VQA Score($\uparrow$) \\
         & OKVQA & VQA on natural images requiring outside knowledge & val & VQA Score($\uparrow$) \\
         & GQA & VQA on scene understanding and reasoning & test-balanced & EM($\uparrow$) \\
         & ScienceQA (Image Set) & Multi-choice VQA on a diverse set of science topics & test & Accuracy($\uparrow$) \\
         & VizWiz & VQA on photos taken by people who are blind & test-dev & VQA Score($\uparrow$)\\
         \midrule
         \multirow{2}{*}{\makecell[l]{Text-oriented \\ VQA}} & TextVQA & VQA on natural images containing text & val & VQA Score($\uparrow$) \\
         & OCRVQA & VQA on images of book covers & test & EM($\uparrow$) \\
         \midrule
         \multirow{3}{*}{\makecell[l]{Refer \\ Expression \\ Comprehension}} & RefCOCO & Refer grounding on natural images & val \& testA \& testB & Accuracy($\uparrow$) \\
          & RefCOCO+ & Refer grounding on natural images & val \& testA \& testB & Accuracy($\uparrow$) \\
          & RefCOCOg & Refer grounding on natural images & val \& test & Accuracy($\uparrow$) \\
         \bottomrule
    \end{tabular}}
        \caption{Summary of the evaluation benchmarks.}
    \label{tab:benchmark}
\vspace{1em}
\end{table*}

\begin{table*}[!h]
\centering
\scalebox{1.09}{
\begin{tabular}{@{}l ccccccccc@{}}
\toprule
\multirow{2}{*}{model} & \multirow{2}{*}{Nocaps} &\multirow{2}{*}{Flickr30k}& \multirow{2}{*}{VQAv2} &   \multirow{2}{*}{OKVQA} & \multirow{2}{*}{GQA} & \multirow{2}{*}{TextVQA} & \multicolumn{2}{c}{RefCOCOg} & \\
&&&&&&&val&test&\\
\midrule
Lyrics & \textbf{126.8} & 85.4 & 81.2 & 58.2 & \textbf{62.4} & 69.4 & \textbf{87.23} & \textbf{88.26}   \\
w/o specific train split & 125.2 & \textbf{86.1} & \textbf{82.5} & \textbf{58.9} & 61.2 & \textbf{69.8} &  86.89 & 88.20  \\
 \bottomrule
\end{tabular}
}
\caption{Results of the ablation experiments on specific train split during the instruction fine-tuning stage.}
\label{tab:ablation_appendix}
\end{table*}

\begin{table}[!t]
    \centering
    \fontsize{8}{11}\selectfont
\caption{Ablation study on model architecture. \vspace{-0.1in}}
\scalebox{1.06}{
    \begin{tabular}{l|cccc}
         \toprule
        Architecture & VQAv2 & RefCOCOg$T$ & LLaVA$^W$ & HallB \\
         \midrule
        \rowcolor[HTML]{F2F3F5}
         \textbf{Lyrics}   &  \textbf{81.2}  &  \textbf{88.26}  &  \textbf{76.9}  &  \textbf{62.6} \\
         w/o ViT  &  78.5  &  86.11  &  74.6  &  60.1 \\
         w/o ODM  &  80.6  &  87.55  &  76.1  &  60.7 \\
         w/o SSM  &  80.8  &  87.08  &  76.5  &  59.2 \\
         w/o VR   &  77.2  &  83.25  &  73.3  &  56.8 \\  
         \bottomrule
    \end{tabular}}
    \label{tab:ablation_model_architecture}
    \vspace{-0.1in}
\end{table}

\begin{table*}[h!]
\centering
\scalebox{0.96}{%
\begin{tabular}{l|l}
\toprule
Task & Instruction Template \\ \midrule
\begin{tabular}[c]{@{}l@{}}Image\\ Captioning\end{tabular} & \begin{tabular}[c]{@{}l@{}} \textless{}Image\textgreater Write a short description for the image.\\ \textless{}Image\textgreater Write a description for the image.\\ \textless{}Image\textgreater Provide a description of what is presented in the photo.\\ \textless{}Image\textgreater Briefly describe the content of the image.\\ \textless{}Image\textgreater Look at the image and describe what you see in a simple and clear manner. \\ \textless{}Image\textgreater Could you use a few words to describe what you perceive in the photo?\\ \textless{}Image\textgreater Please provide a short depiction of the picture.\\ \textless{}Image\textgreater Summarize what this image depicts in a simple and concise manner.\\ \textless{}Image\textgreater Provide a simple and clear description of the image, suitable for all audiences. \end{tabular} \\ \midrule

\begin{tabular}[c]{@{}l@{}}Visual\\ Question \\ Answering \end{tabular} & \begin{tabular}[c]{@{}l@{}}\textless{}Image\textgreater \{Question\}\\ \textless{}Image\textgreater Question: \{Question\}\\ \textless{}Image\textgreater Question: \{Question\} Answer:\\ \textless{}Image\textgreater Given the image, answer the following question: \{Question\} \\ \textless{}Image\textgreater With the aid of the following image, offer a straightforward, short response to: \{Question\}.\\ \textless{}Image\textgreater Based on the image, respond to this question with a short answer: \{Question\}. Answer:\\ \textless{}Image\textgreater Use the provided image to answer the question as short as possible: \{Question\}\\ \textless{}Image\textgreater What is the answer to the following question? \{Question\}\\ \textless{}Image\textgreater Refer to the information in the image to provide a minimalist answer to: \{Question\} \end{tabular} \\ \midrule

\begin{tabular}[c]{@{}l@{}} Text-oriented \\ Visual\\ Question \\ Answering \end{tabular} & \begin{tabular}[c]{@{}l@{}}\textless{}Image\textgreater Question: \{Question\}\\ \textless{}Image\textgreater Question: \{Question\} Answer:\\ \textless{}Image\textgreater Analyze the textual content in this image and provide a short answer to: \{Question\}. \\ \textless{}Image\textgreater Look at the text in the image provided and succinctly answer: \{Question\}.\\ \textless{}Image\textgreater With the help of text in the following image, offer a simple, short response to: \{Question\}.\\ \textless{}Image\textgreater Refer to the textual data in the image to provide a brief answer to: \{Question\}. \\ \end{tabular} \\ \midrule

\begin{tabular}[c]{@{}l@{}} Grounded \\ Captioning \end{tabular} & \begin{tabular}[c]{@{}l@{}}\textless{}Image\textgreater Write a description for the target object in the image. \\ \textless{}Image\textgreater Provide a short caption focusing on the highlighted object in this image. \\ \textless{}Image\textgreater Describe the specific object indicated in the following image, keeping the description brief. \\ \textless{}Image\textgreater Explain what the object marked in the image is, using a concise description. \\ \textless{}Image\textgreater Identify and describe the key object in this image, using a short and clear description. \\ \end{tabular} \\ \midrule

\begin{tabular}[c]{@{}l@{}} Referring \\ Expression \\ Comprehension \end{tabular} & \begin{tabular}[c]{@{}l@{}}\textless{}Image\textgreater In the given image, could you find and tell me the coordinates of \{$Tag$\}? \\ \textless{}Image\textgreater In the coordinate \{$Bbox$\} of the image, can you observe the object \{$Tag$\}. \\ \textless{}Image\textgreater Locate the \{$Tag$\} in this image and provide a brief description of its position. \\ \textless{}Image\textgreater Confirm the presence of \{$Tag$\} in the bounding box \{$Bbox$\} in the image. \\
\textless{}Image\textgreater Search for \{$Tag$\} in the image and give its coordinates if found. \\
\textless{}Image\textgreater Can you find the spatial location or coordinates of \{$Tag$\} in the image shown here? 
 \\ \end{tabular}
\\ \midrule

\begin{tabular}[c]{@{}l@{}} Referential \\ Dialogue \end{tabular} & \begin{tabular}[c]{@{}l@{}}\textless{}Image\textgreater Focus on the object \{$Tag\ \& \ Bbox$\} in the image, and answer the question: \{Question\}. \\ 
\textless{}Image\textgreater Could you provide a descriptive caption for the object \{$Tag\ \& \ Bbox$\} in the image? \\ \textless{}Image\textgreater Regarding the object specified as \{$Tag\ \& \ Bbox$\}, please respond to: \{Question\}. \\ \textless{}Image\textgreater Explain the features or details of the object identified by \{$Tag\ \& \ Bbox$\} in the image. \\
\textless{}Image\textgreater Create a caption that describes the area or object marked as \{$Tag\ \& \ Bbox$\} in the image. \\
\textless{}Image\textgreater Refer to the object \{$Tag\ \& \ Bbox$\} in the image, and provide an answer to: \{Question\}. \\
\end{tabular}
\\ \midrule

\begin{tabular}[c]{@{}l@{}} Multi-choice \\ Visual\\ Question \\ Answering  \end{tabular} & \begin{tabular}[c]{@{}l@{}}\textless{}Image\textgreater Question: \{Question\} Options: \{Option\}. Answer: \\ 
\textless{}Image\textgreater For the question: \{Question\}, choose the most suitable answer from options: \{Option\}. \\ \textless{}Image\textgreater Examine the image and answer the question: \{Question\}. Your choices are: \{Option\}. \\ \textless{}Image\textgreater Respond to the question: \{Question\} among options: \{Option\}, select your response: \\
\textless{}Image\textgreater Consider the question: \{Question\} and options: \{Option\}. Please provide your answer: \\
\end{tabular}
\\

\bottomrule
\end{tabular}%
}
\vspace{2pt}
\caption{Instruction templates used for transforming the conventional vision-language datasets into instruction tuning data.}
\label{tab:appendix_instruct_template}
\end{table*}

\begin{figure*}[b]
    \centering
    \includegraphics[width=1\linewidth, trim=0 10 0 0]{figures/graph_result.pdf}
    \caption{Our Lyrics achieves state-of-the-art performance on a broad range of vision-language tasks compared with other generalist models.}
    \label{fig:graph_result}
    \vspace{-0.2in}
\end{figure*}

\clearpage

{
    \small
    \bibliographystyle{ieeenat_fullname}
    \bibliography{main}
}